\documentclass[11pt]{article}

\usepackage[preprint]{acl}

\usepackage{times}
\usepackage{latexsym}

\usepackage[T1]{fontenc}

\usepackage[utf8]{inputenc}

\usepackage{microtype}

\usepackage{inconsolata}

\usepackage{graphicx}

\usepackage{amsmath}
\usepackage{amssymb}

\usepackage{booktabs}
\usepackage{multirow}
\usepackage{array}
\usepackage{tabularx}

\title{Semantic Gradients Interactions in SSD: A Case Study in Racial Identity and Hate Speech}

\author{Felix Ostrowicki \\
  Independent Researcher \\
  \texttt{fx.ostrowski@gmail.com} \\\And
  Hubert Plisiecki \\
  IDEAS Research Institute \\
  \texttt{hplisiecki@gmail.com} \\}

\begin{document}
\maketitle

\begin{abstract}
We introduce interaction SSD, an extension of Supervised Semantic Differential
that models how semantic meaning varies across moderators such as groups,
traits, or conditions making this variation testable and interpretable.
The method estimates a main semantic gradient, an
interaction gradient, and conditional gradients, all interpretable through
standard SSD tools. We illustrate it on the UC Berkeley Measuring Hate Speech
corpus, testing whether annotator racial identity moderates hate-speech
judgments of comments targeting people of color. The interaction model detects
a significant moderation effect: the shared gradient contrasts dehumanizing
hostility with counter-speech, while the interaction gradient reveals smaller
group-linked differences in which semantic cues predict hate-speech ratings.
Interaction SSD makes moderated meaning--outcome relationships statistically
testable and interpretable.
\end{abstract}

\noindent\textbf{Content warning:} This paper contains examples of hate
speech, slurs, and other offensive language included for research and
analytical purposes. Slurs are redacted using a partial-asterisk
convention (first and last letters retained, middle characters replaced):
e.g., \textit{n****r}, \textit{f****t}.

\section{Introduction}
\label{sec:intro}

Supervised Semantic Differential (SSD) is a recent mixed quantitative--interpretive
method for analyzing how the meaning of concepts and texts varies with external
variables such as traits, attitudes, or judgments \citep{plisiecki2025}. SSD represents texts or concept-specific contexts as vectors in an embedding space, regresses outcome variables onto these representations, tracing a semantic gradient in the embedding space which describes the semantic shift of the concept under study. The positive and negative poles of this gradient can then be explored through nearest-neighbor retrieval, clustering, and representative text snippets, allowing researchers to interpret statistical associations as structured differences in meaning.

However, the canonical SSD model only estimates a single semantic gradient for the
full sample. This is appropriate when the goal is to recover the overall
semantic structure associated with an outcome, but many psychological and
social-scientific questions are concerned with whether this structure differs across groups, identities, contexts, or conditions. For example, the same textual
content may carry different evaluative implications depending on who judges it,
and the same semantic cues may be more or less predictive of an outcome across
theoretically meaningful subgroups. In such cases, a single SSD gradient can
average over heterogeneous meaning--outcome relationships, and obscure the
moderation effects that are central to the research question.

This work introduces an interaction extension of SSD that allows semantic
gradients to vary as a function of a moderator, yielding a main semantic
gradient, an interaction gradient, and conditional gradients at substantively
meaningful values of the moderator, all interpretable with standard SSD tools.

We showcase the method using the Berkeley Measuring Hate Speech corpus
\citep{kennedy2020constructing}, a large-scale annotation dataset containing online text samples, hate speech ratings, information about the targets of speech, and annotator demographics. Specifically, we examine hate speech judgments for texts aimed at people of color and ask whether annotator racial identity,
operationalized as a white / people of color binary split, moderates the
semantic structure of these judgments. The case study serves primarily as an
illustration of the extended method; the main contribution is methodological.
We show how SSD can be extended from estimating a single interpretable semantic
gradient to modeling semantic moderation effects, enabling transparent analysis
of how meaning--judgment relationships differ across socially meaningful
contexts.

\section{Background: Supervised Semantic Differential}
\label{sec:background}

SSD assumes a collection of documents $d_i$ paired with continuous outcomes
$y_i$. In its canonical formulation, SSD uses a lexicon denoting a concept of
interest to aggregate the local semantic contexts in which that concept appears,
yielding Personal Concept Vectors (PCVs) that represent how the concept is used
in each author's texts. In whole-document mode, each text is treated as the
semantic unit of analysis \citep{plisiecki2025}.

Each document or concept-specific context is mapped to a dense vector
$x_i \in \mathbb{R}^D$ using a fixed embedding model, here with SIF-weighted
averaging \citep{arora2017} and removal of the top principal component
to reduce anisotropy \citep{mu2017}.
Because the original embedding space is high-dimensional and partly redundant,
SSD applies PCA \citep{jolliffe2002} before regression. The number of retained components is selected
with a PCA sweep, which evaluates candidate dimensionalities $K$ in terms of
retained variance, coherence of the gradient's local semantic neighborhoods
\citep{mimno2011}, and stability across nearby values of $K$ \citep{plisiecki2026}. This yields
a reduced representation $\tilde{x}_i \in \mathbb{R}^K$ used in the standard
SSD regression:
\[
y_i = \alpha + \beta^\top \tilde{x}_i + \epsilon_i .
\]

The coefficients $\beta$ are back-projected into the original embedding space
to obtain an interpretable semantic gradient $\hat{\beta}$. SSD interprets the
positive and negative poles of this gradient by retrieving nearest neighbors,
clustering them separately, and extracting original text snippets aligned with
the cluster centroids. The method therefore combines statistical
estimation with traceable semantic interpretation: the regression tests whether
semantic variation is associated with the outcome, while the back-projected
gradient, lexical neighborhoods, and snippets characterize the form of that
association.

\section{Interaction SSD}
\label{sec:interaction}

We introduce an interaction extension of SSD for cases where the
meaning--outcome relationship is expected to vary with a moderator, such as a
group indicator, individual-difference variable, or experimental condition.
After the standard SSD embedding pipeline and PCA sweep, each document is
represented by a selected $K$-dimensional vector
$\tilde{x}_i \in \mathbb{R}^K$. The outcome $y_i$ and moderator $m_i$ are
standardized before regression.

The model augments standard SSD with a moderator main effect and a semantic
interaction block:
\begin{align*}
y_i &= \alpha
+ \beta^\top \tilde{x}_i
+ \gamma m_i
+ \delta^\top(\tilde{x}_i m_i)
+ \epsilon_i .
\end{align*}
Here, $\beta$ captures the baseline semantic association with the outcome,
$\gamma$ captures the non-semantic main effect of the moderator, and $\delta$
tests whether the semantic association itself changes as a function of the
moderator. The semantic and interaction blocks are evaluated with Wald $F$
tests over their respective $K$ coefficients, while $\gamma$ is evaluated as a
single regression coefficient. A full matrix-form specification, including the
OLS estimator, covariance matrix, block tests, and back-projection equations, is
provided in Appendix~\ref{appendix:formal}.

\begin{table*}[t]
\centering
\small
\renewcommand{\arraystretch}{1.15}
\begin{tabular}{lrrrrr}
\toprule
\textbf{Model term} 
& \textbf{Estimate} 
& \textbf{SE} 
& \textbf{Test} 
& \textbf{df} 
& \textbf{Effect size} \\
\midrule
Overall model 
& --- 
& --- 
& $F = 702.78$ 
& $125, 44687$ 
& $R^2 = .663$, $R^2_{\mathrm{adj}} = .662$ \\

Semantic block $\boldsymbol{\beta}$ 
& --- 
& --- 
& $F = 1231.21$ 
& $62, 44687$ 
& --- \\

Race main effect $\hat{\gamma}$ 
& $.028$ 
& $.003$ 
& $t = 8.87$ 
& $44687$ 
& --- \\

Interaction block $\boldsymbol{\delta}$ 
& --- 
& --- 
& $F = 13.76$ 
& $62, 44687$ 
& partial $R^2 = .019$ \\
\bottomrule
\end{tabular}
\caption{Interaction-SSD regression results. $n = 44{,}813$; outcome and IV
were $z$-scored; $K = 62$. Partial $R^2$ is reported for the interaction block
and represents the proportion of residual variance explained by the interaction
after accounting for the shared semantic rating structure and the race main
effect.}
\label{tab:cs-regression}
\end{table*}

As in standard SSD, coefficient vectors are back-projected from PCA space into
the original embedding space. The baseline and interaction blocks yield
\[
g_{\mathrm{main}} = P\beta,
\qquad
g_{\mathrm{int}} = P\delta ,
\]
where $P$ denotes the linear back-projection from the selected PCA space. The
main gradient describes the shared semantic direction associated with higher
outcome values, whereas the interaction gradient describes how that direction
changes with the moderator.

The model also yields conditional gradients analogous to simple slopes in
moderated regression \citep{aiken1991}:
\[
g(m^\ast) = g_{\mathrm{main}} + m^\ast g_{\mathrm{int}} .
\]
For continuous moderators, $m^\ast$ can be set to values such as one standard
deviation below and above the mean; for binary moderators, it can be evaluated
at the standardized group values. The main, interaction, and conditional
gradients are then interpreted with the usual SSD nearest-neighbor, clustering,
and snippet tools. Thus, the regression tests whether moderation is present,
while the back-projected gradients make the semantic form of that moderation
inspectable.

\section{Case Study: Moderated Semantic Gradients in Hate-Speech Judgments}
\label{sec:casestudy}

We applied interaction-moderated SSD to $n = 44{,}813$ annotation records
from the UC Berkeley Measuring Hate Speech corpus
\citep{kennedy2020constructing}, restricted to comments whose target was coded as people of color, using annotator race as a binary moderator
(POC\,=\,0, white\,=\,1, then standardized) following \citet{sachdeva2022}.
We refer to group-linked differences in the semantic cues associated with
hate-speech ratings as \textit{annotator identity sensitivity}
\citep{sachdeva2022}. Full corpus and preprocessing details are in
Appendix~\ref{appendix:data}.

Texts were embedded as whole documents using 300-dimensional Common Crawl
GloVe vectors \citep{pennington2014glove} with SIF weighting
\citep{arora2017}. Because the corpus
spans heterogeneous targets and rhetorical registers, we decided to focus on the whole texts as unit of analysis, instead of creating a specific concept lexicon. The interaction model was fit at $K = 62$ PCA components, selected by the PCA sweep over
$K \in \{2, 4, 6, \ldots, 122\}$ \citep{plisiecki2025}.

Table~\ref{tab:cs-regression} summarizes the regression results. The
model explained substantial variance in hate-speech scores
($R^{2}_{\mathrm{adj}} = .662$). The semantic block was strongly
significant, as was the annotator-race main effect: white annotators
assigned slightly higher hate-speech scores on average
($\hat{\gamma} = .028$, $SE = .003$, $t = 8.87$, $p < 10^{-16}$).
Critically, the interaction block was also significant
($F(62, 44687) = 13.76$, $p < 10^{-16}$), indicating that the semantic
axis predictive of hate-speech judgments differed by annotator race.
The corresponding partial $R^2$ was .019, meaning that the interaction
explained approximately 1.9\% of the residual variance left after
accounting for the shared semantic rating structure and the race main
effect. Thus, the moderation effect was statistically robust but small:
it should be interpreted as a reliable second-order deviation from a
largely shared semantic structure rather than as evidence for entirely
separate rating systems.

\begin{table*}[t]
\centering
\footnotesize
\renewcommand{\arraystretch}{1.03}
\setlength{\tabcolsep}{2.5pt}

\begin{tabularx}{\textwidth}{
@{}c r >{\raggedright\arraybackslash}X
@{\hspace{0.7em}}
c r >{\raggedright\arraybackslash}X@{}}
\toprule
\multicolumn{3}{c}{\textbf{Main gradient $\hat{g}_{\mathrm{m}}$}} &
\multicolumn{3}{c}{\textbf{Interaction gradient $\hat{g}_{\mathrm{i}}$}} \\
\cmidrule(r{0.7em}){1-3}
\cmidrule(l{0.7em}){4-6}
\textbf{Pole} & \textbf{$N$} & \textbf{Theme} &
\textbf{Pole} & \textbf{$N$} & \textbf{Theme} \\
\midrule
$+$ & 33 & Slur contempt: \textit{bastard, n****r, f****t, monkey}
&
$+$ & 46 & Attack/violence: \textit{attack, kill, sniper, shooter}
\\
$+$ & 31 & Eliminationist: \textit{terrorists, nazis, kill, murderers}
&
$+$ & 11 & Contagion framing: \textit{swine, meningitis, leprosy}
\\
$+$ & 36 & Vermin: \textit{rodents, fleas, cockroaches}
&
$+$ & 43 & Meta-discursive racism: \textit{racism, bigotry, accusation}
\\
$-$ & 25 & Diasporic civic talk: \textit{city names}
&
$-$ & 64 & Affective racial commentary: \textit{electrifying, vibrancy}
\\
$-$ & 47 & Reflexive race talk: \textit{nuances, dynamism, attuned}
&
$-$ & 36 & Spanish intra-community: \textit{nuevo, m\'{u}sica, barrio}
\\
$-$ & 28 & POC achievement: \textit{acclaimed, songwriter, performer}
&
& &
\\
\bottomrule
\end{tabularx}

\caption{Main and interaction gradient clusters in hate-speech annotations. The main gradient captures the semantic structure shared across annotators, whereas the interaction gradient captures race-moderated deviations from that shared structure.}
\label{tab:cs-clusters}
\end{table*}

To interpret the model qualitatively, we clustered nearest neighbors on the positive and negative poles of the main, conditional, and interaction gradients.
Table~\ref{tab:cs-clusters} summarizes the principal themes for the
main effect axis ($\hat{g}_{\mathrm{main}}$) and the interaction axis
($\hat{g}_{\mathrm{int}}$); full per-gradient tables with representative
excerpts are reported in Appendix~\ref{appendix:ssd-clusters}. The average axis $\hat{g}_{\mathrm{main}}$ recovered a coherent contrast
between overtly hostile discourse and community-affirming counter-speech.
Its positive pole grouped slur-based contempt, eliminationist
categorization, and vermin metaphors; its negative pole grouped
diasporic civic discourse, reflexive engagement with race, and
celebratory recognition of POC cultural achievement. The conditional
gradients for white and POC annotators showed broadly similar high-score
poles but differed in their local organization, as detailed in
Appendix~\ref{appendix:ssd-clusters}.

The interaction axis $\hat{g}_{\mathrm{int}}$ marks where the two
annotator groups diverged from this shared structure. Its positive pole,
corresponding to semantics to which white annotators were uniquely more sensitive, was organized around explicit and surface-marked hostility: attack
vocabulary, biomedical/contagion framings, and meta-discursive racism
terms. Its negative pole, corresponding to uniquely higher
POC-annotator sensitivity, contained more context-dependent material,
including affective racial commentary and Spanish-language
intra-community conflict. Appendix~\ref{app:cc840b} reports a
robustness reanalysis with the Common Crawl 840B GloVe model: the
statistical moderation finding replicates, while local gradient
neighborhoods vary with embedding geometry. The code used to conduct the case study is publicly available on OSF \citep{ostrowicki2026code}.

\section{Discussion and Conclusion}
\label{sec:discussion}

Interaction SSD extends Supervised Semantic Differential to moderated
meaning--outcome relationships, making them statistically testable and
semantically interpretable across groups, traits, or conditions. The model
estimates a shared semantic gradient, an interaction gradient, and conditional
gradients, all of which can be back-projected into the original embedding space
and interpreted with standard SSD tools.

The case-study results are consistent with prior work showing that in-group membership heightens sensitivity to hate speech targeting one's own community
\citep{operario2001}, and that annotator race predicts systematic labeling
differences \citep{sachdeva2022,giorgi2025}. Overall, the model recovered a
largely shared rating structure, contrasting slur-based contempt, eliminationist
language, and dehumanizing metaphors with civic, reflexive, and
community-affirming discourse. The significant interaction block indicated that
this shared gradient was modestly moderated by racial group association, with
group-linked differences emerging mainly at the margins of the rating scale.
Thus, the interaction should be read as a small but reliable deviation from a
dominant shared axis, not as evidence of fundamentally different evaluative
frameworks.

In addition, this second-order character also limits how strongly the qualitative clusters should be interpreted. Because the gradient is estimated from differences between conditional meaning--rating relationships, its local
nearest-neighbor structure is expected to be less stable than the main gradient. The robustness analysis in Appendix~\ref{app:cc840b} supports this interpretation: the statistical moderation effect replicated across embedding models, while the semantic neighborhoods around the interaction gradient changed significantly. Interaction SSD is therefore best used as an exploratory and theory-building tool, unless applied to much stronger effects: it can identify where moderated semantic associations may occur and generate hypotheses for more targeted follow-up studies.

Future work can extend this framework beyond single-outcome moderated regression through integration with structural equation modeling \citep{plisiecki2026}, where semantic gradients could enter as indicators,
predictors, or outcomes in larger latent-variable systems. This would
allow SSD-based semantic measurement to be linked to
multi-group models, and measurement-error-aware theories of attitudes,
identities, judgments, and behavior.

\section*{Limitations}
\label{sec:limitations}

The analysis treats annotation records as independent observations and does
not model the crossed random effects of annotators and comments; standard
errors and F-statistics should therefore be read as model-based descriptive
evidence rather than as output from a full mixed-effects analysis. The binary
white/POC moderator is a coarse proxy for social positioning: annotator-group
differences conflate within-group heterogeneity and should not be generalized
to specific target--annotator combinations beyond those represented in this
corpus. More broadly, results are specific to the UC Berkeley Measuring Hate
Speech corpus---English-language social media text, US-based MTurk
annotators---and may not transfer to other annotation pipelines, languages, or
target populations. The interaction SSD model addresses moderation at the level
of a single binary contrast; extending the method to continuous or multiway
moderators, and to corpora with more structured sampling of annotator identity,
remains an open direction. Finally, as the CC840B reanalysis in
Appendix~\ref{app:cc840b} illustrates, the semantic interpretation of the
interaction gradient is contingent on the choice of embedding model; gradient
neighborhoods should be treated as exploratory maps of one model's geometry
rather than as stable empirical facts about annotator behaviour.

\section*{Ethical Considerations}

This paper analyzes hate speech annotations paired with annotator demographic
data, raising issues of content sensitivity, identity representation, and
interpretive risk. The corpus contains slurs and graphic offensive language
targeting racial and other minority groups; all displayed examples follow the
redaction policy described at the outset of the paper, and a content warning
is provided before the main text. The binary racial coding used as a moderator
operationalizes a socially constructed category as a fixed two-level variable;
this is a deliberate analytical simplification, not a claim about the stability
or boundedness of racial identity. Findings describe group-level statistical
patterns in semantic gradients and should not be used to characterize
individual annotators, to argue against demographic diversity in annotation, or
to essentialize differences between racial groups. Interaction SSD is a tool
for surfacing and interrogating meaning--outcome relationships, not a profiling
or classification technology; the gradients it recovers are low-norm,
second-order signals that support interpretive exploration rather than
predictive inference. Portions of this manuscript were drafted or revised with
the assistance of a large language model; all such passages were reviewed,
validated, and where necessary corrected by the authors to ensure accuracy and
faithfulness to the data.

\bibliography{references}

\appendix

\section{Hate Speech Dataset Details}
\label{appendix:data}

The UC Berkeley Measuring Hate Speech corpus
\citep{kennedy2020constructing} is a large-scale annotation dataset
designed to measure hate speech as a continuous construct. Comments were
collected via public APIs from YouTube, Twitter, and Reddit and filtered
to English-language posts between 4 and 600 characters. The final corpus
comprised approximately 50,000 comments, each reviewed by a mean of
approximately four independent annotators recruited via Amazon Mechanical
Turk. In total, the original corpus included approximately 10,000 workers
resident in the United States; the median task duration was 49 minutes and
the median compensation was \$8.57 per hour.

Each annotator completed a structured labeling instrument covering ten
ordinal sub-dimensions per comment, together with a binary hate-speech
judgment. A continuous \texttt{hate\_speech\_score} was derived from the
ordinal items using a faceted many-facet Rasch item response theory (IRT)
model, which jointly estimates comment severity, item difficulty, and
individual rater leniency \citep{kennedy2020constructing}. The resulting
scale adjusts for annotator severity bias and ranges from strong
counter-speech at the negative pole to genocidal hate speech at the
positive pole.

Annotators also provided self-reported demographic information, including
race/ethnicity. Following \citet{sachdeva2022}, we operationalized
annotator race as a binary white/POC contrast. This coding was used as a
coarse proxy for differential social positioning relative to the
racialized targets most commonly referenced in the corpus; it should not
be interpreted as a fine-grained model of racial identity or as a
target-specific measure of in-group membership. We therefore interpret
race-linked interaction effects as differences in which semantic cues are
associated with higher hate-speech ratings, not as stable properties of
racial groups or individual annotators. In the main text, we refer to
this pattern as \textit{annotator identity sensitivity}.

 Prior to analysis, the corpus was filtered to annotation records for comments targeting people of color, as identified by the target-group coding in the original dataset. This restriction focuses the moderation analysis on a context where annotator in-group/out-group status is well-defined relative to the speech target. For the present analysis, annotations were categorized into one of two groups:
white and people of color (POC). The binary moderator was coded as
POC\,=\,0 and white\,=\,1 before standardization. After excluding records
with missing annotator race or outcome data, the analytic sample comprised
$n = 44{,}813$ annotation records. White annotators constituted
approximately 75\% of the analytic sample
($M_{\mathrm{IV}} = 0.75$, $SD = 0.43$). The
\texttt{hate\_speech\_score} had mean $0.06$ ($SD = 2.31$; range
$[-8.21, 6.30]$), consistent with the overall corpus distribution
reported by \citet{kennedy2020constructing}. White annotators assigned
slightly higher scores on average ($M = 0.07$, $SD = 2.31$,
$n = 32{,}900$) than non-white annotators ($M = 0.03$, $SD = 2.33$,
$n = 8{,}861$).

Tokenization and linguistic preprocessing were performed using spaCy~3.x
\citep{montani2023} with the \texttt{en\_core\_web\_lg} English pipeline. Standard English
stopwords were removed prior to embedding. Documents were embedded using
the 300-dimensional GloVe model trained on 42 billion Common Crawl tokens
\citep{pennington2014glove} with SIF weighting
($a = 10^{-3}$; \citealp{arora2017}). Word vectors were
$L^2$-normalized to reduce the frequency--norm correlation
\citep{wilson2015}, and the top principal component of the document space
was removed following the all-but-the-top correction of \citet{mu2017}.

No hate-speech lexicon was used to anchor the analysis. The corpus spans
heterogeneous targets, topics, and rhetorical registers, making a single
target lexicon inappropriate for the present demonstration. Because all
comments were drawn from a shared annotation task, each comment was
instead treated as a whole-document semantic unit. The interaction model
was fit at $K = 62$ PCA components selected by the joint
interpretability--stability (AUCK) sweep over
$K \in \{2, 12, 22, \ldots, 122\}$; remaining settings followed the
original SSD configuration \citep{plisiecki2025}.

The UC Berkeley Measuring Hate Speech corpus is released under the Creative
Commons Attribution 4.0 license (CC-BY 4.0). The GloVe pre-trained word
vectors are distributed under the Public Domain Dedication and License v1.0
(PDDL). The spaCy NLP library is released under the MIT License. All three
artifacts were used for non-commercial research purposes consistent with their
respective licenses.

The present study is a secondary analysis of a publicly available, previously
published corpus. No new participants were recruited and no new data were
collected; participant consent was the responsibility of the original data
collectors.

The present work involves secondary analysis only and did not require
independent ethics board review. Details of the original corpus collection
protocol, including annotator compensation and data handling, are reported in
\citet{kennedy2020constructing}.

\section{Corpus and Lexicon-Specific Semantic Results}
\label{appendix:ssd-clusters}

\paragraph{Interpretive status of cluster labels.}
The cluster labels reported below are interpretive summaries of local
nearest-neighbor neighborhoods around the recovered SSD gradients. They
should be read as qualitative aids for understanding the fitted semantic
directions, not as discrete latent classes or independently validated
categories. The reported themes therefore describe how the embedding
model organizes neighborhoods around each gradient pole, rather than
directly identifying psychological mechanisms in annotators.

For the conditional gradients, positive-pole neighborhoods indicate
semantic regions associated with higher predicted hate-speech scores
within a given annotator group, whereas negative-pole neighborhoods
indicate regions associated with lower scores or counter-speech. For the
interaction gradient, the interpretation is different: positive and
negative poles indicate relative group differences in the strength of the
meaning--rating association. Thus, $\hat{g}_{\mathrm{int}}$ should be
read as a divergence axis, not as an absolute hate-speech axis for either
group.

The following sections report the corpus- and lexicon-specific semantic
cluster results for each of the three analytic groups derived from the
Supervised Semantic Differential (SSD) analysis: the general tendency
gradient estimated without conditioning on annotator race
(\textit{g}\textsubscript{main}), the conditional gradient for white
annotators (\textit{g}\textsubscript{high}), and the conditional gradient
for POC annotators (\textit{g}\textsubscript{low}). For each group,
clusters are described in terms of their semantic themes at the positive
(high hate-speech score) and negative (low hate-speech score /
counter-speech) poles. Sub-dimension labels follow
\citet{kennedy2020constructing}.

\subsection{General Tendency (\textit{g}\textsubscript{main})}

\subsubsection*{Gradient Summary}
The full corpus yielded a well-structured semantic gradient capturing a
broad contrast between overtly hostile, dehumanizing racial discourse and
reflective, community-affirming counter-speech, or relatively neutral
statements. Across annotators without race stratification, the recovered axis
ran from language combining explicit slur use, exterminatory threat, and
animalization at the high end to geographically grounded, metacognitive, and
celebratory discourse at the low end intermixed with neutral statements. The
gradient accounts for substantial variance in hate speech scores, and both
poles were populated by lexically coherent, thematically interpretable
clusters.

\subsubsection*{Positive Pole}
The positive pole was organized around three co-activating rhetorical
registers. One cluster centered on explicit slur-based derogation and coarse
contempt vocabulary directed at racial and sexual minority groups. A second
emphasized armed hostility, exterminatory intent, and political
extremism---vocabulary associated with calling for violence and genocide as
defined by \citet{kennedy2020constructing}. A third reflected animalistic
dehumanization, specifically the vermin and infestation sub-register
historically linked to genocidal propaganda \citep{stanton1998}. Taken together, these clusters
suggest that the corpus-level hate speech signal is strongest when explicit
degradation, physical threat, and subhuman framing appear in semantic
proximity---consistent with \citeauthor{kennedy2020constructing}'s finding
that the \textit{dehumanize}, \textit{violence}, and \textit{genocide}
sub-dimensions load together at the high end of the composite hate speech
score.

\subsubsection*{Negative Pole}
The negative pole was organized around three distinct counter-speech modes.
One cluster was anchored by geographically specific, situationally grounded
discourse referencing diasporic and multicultural community life across
multiple continents. A second emphasized reflexive, metacognitive engagement
with race and racism---language oriented toward nuance, historical awareness,
and deliberate sensitivity. A third foregrounded celebratory recognition of
Black and other POC's cultural achievement in artistic and public life. Taken
together, these clusters suggest that lower hate speech scores were
associated with discourse that actively raises group status, defends targeted
communities, and participates in recognizable counter-speech
practices---mapping onto the \textit{respect}, \textit{status}, and
\textit{attack\_defend} dimensions of \citeauthor{kennedy2020constructing}'s
(\citeyear{kennedy2020constructing}) annotation scheme.

\subsubsection*{Interpretation Note}
This corpus illustrates a relatively clear structural contrast between
language that degrades and threatens and language that affirms and defends.
The nearest-neighbor clusters and representative excerpts were broadly
convergent at both poles, with the positive pole showing particularly high
cosine alignment with canonical examples of slur use, threat, and
dehumanization. One point of interpretive complexity is that the vermin
cluster captures both primary dehumanizing texts and counter-speech that
deploys the same vocabulary to condemn racists---a polarity ambiguity that
the IRT-derived hate speech score partially resolves through annotator
consensus but that warrants attention in downstream interpretation.

\begin{table}[htbp]
\centering
\small
\begin{tabular}{cc>{\raggedright\arraybackslash}p{0.68\linewidth}}
\toprule
Pole & Size & Summary (Top Words / Excerpt) \\
\midrule
$+$ & 33 & \emph{Slur contempt}: \textit{bastard, n****r, f****t, monkey, pig} --- ``She is a fucking dumb monkey'' \\
$+$ & 31 & \emph{Eliminationist}: \textit{terrorists, criminals, nazis, kill, murderers} --- ``Other countries say the same thing about YOU EVIL LAND INVADING MURDEROUS SAVAGES\ldots'' \\
$+$ & 36 & \emph{Vermin}: \textit{rodents, fleas, parasites, cockroaches, pests} --- ``The usa does not need any more rats and cockroaches, get them out!'' \\
\midrule
$-$ & 25 & \emph{Diasporic civic talk}: \textit{saskatoon, oakville, guelph, moncton, rotorua} --- ``How come? I just recently moved to Klang Valley from Penang\ldots'' \\
$-$ & 47 & \emph{Reflexive race talk}: \textit{nuances, reflecting, dynamism, subtleties, attuned} --- ``Whilst I agree that I've always felt inherently safe and struggle to empathize with those who don't\ldots'' \\
$-$ & 28 & \emph{POC achievement}: \textit{acclaimed, songwriter, showcased, performer, songwriting} --- ``I am honored to be named One of The 100 Most Influential African Women in 2019.'' \\
\bottomrule
\end{tabular}
\caption{Cluster summary for the general tendency gradient ($\hat{g}_{\mathrm{main}}$). Positive pole: high hate-speech score; negative pole: counter-speech / low hate-speech score.}
\label{tab:gmain_pos}
\end{table}

\subsection{POC Annotators (\textit{g}\textsubscript{low})}

\subsubsection*{Gradient Summary}
The POC annotator gradient produced a condensed but high-coherence positive
pole and a structurally elaborated, three-part negative pole. The recovered
contrast ran from a semantically unified field of hostile group
categorization and dehumanization at the high end to a differentiated
counter-speech space encompassing diasporic community discourse, critical
race reflexivity, and celebratory affirmation of POC cultural life at the
low end. The gradient's defining structural feature is its asymmetry in the
opposite direction from the white gradient: fewer, broader clusters at the
hate-speech pole and more, distinct clusters at the counter-speech pole.

\subsubsection*{Positive Pole}
The positive pole was organized around two large, coherent clusters. One
merged criminalization, political extremism, and eliminationist
vocabulary---treating the co-occurrence of hostile group labeling and calls
for annihilation as a single semantic unit. The second clustered the full
vermin and infestation register of zoological dehumanization, achieving the
highest coherence of any positive-pole cluster across all three gradients.
Taken together, these clusters suggest that POC annotators apply a schema
in which the \citet{kennedy2020constructing} \textit{dehumanize} and
\textit{genocide} constructs are treated as co-implicating: the
hate-speech significance of dehumanizing language is understood to derive
not from its surface lexical form but from its combination with hostile
group categorization and implied eliminationist logic. The consolidation
into two clusters rather than four reflects a more structurally integrated
rather than lexically granular detection schema.

\subsubsection*{Negative Pole}
The negative pole was organized around three semantically distinct regions.
One was anchored by non-Western diasporic geolocation, spanning South
Asian, Oceanic, and Canadian city names alongside discourse embedded in
regional politics and multicultural civic life---a substantially broader
geographic imagination than appears in the white annotator gradient. A
second cluster captured critical race reflexivity: deliberate,
nuance-oriented discourse about colonialism, privilege, structural racism,
and cross-cultural empathy. A third centered on celebratory recognition of
Black and POC achievement in artistic and public domains. Taken together,
these clusters suggest that POC annotators encode the
\textit{attack\_defend}, \textit{status}, and \textit{respect} dimensions
of \citet{kennedy2020constructing} as structurally distinct
counter-speech modes---not a single affirmative residual---with language
that actively defends, raises status, and affirms group identity functioning
as three separable semantic anchors of the non-hateful end of the
continuum.

\subsubsection*{Interpretation Note}
This corpus illustrates a structurally richer counter-speech schema than is
found in either the general or white annotator gradients. The
nearest-neighbor clusters and representative excerpts were convergent and
mutually reinforcing at both poles, with particular coherence at the
negative pole where each of the three clusters was populated by thematically
consistent and semantically distinct example texts. The most consequential
interpretive feature is that POC annotators mark celebratory Black and POC
cultural discourse as a semantically \textit{specific} counter-speech
anchor---not merely the absence of hostility---suggesting a gradient
organized around a meaningful social opposition rather than a simple
threat-detection threshold.

\begin{table}[htbp]
\centering
\small
\begin{tabular}{cc>{\raggedright\arraybackslash}p{0.68\linewidth}}
\toprule
Pole & Size & Summary (Top Words / Excerpt) \\
\midrule
$+$ & 55 & \emph{Eliminationist/dehumanization}: \textit{bastards, criminals, terrorists, kill, nazis} --- ``Other countries say the same thing about YOU EVIL LAND INVADING MURDEROUS SAVAGES\ldots'' \\
$+$ & 45 & \emph{Zoological dehumanization}: \textit{rodents, cockroaches, fleas, worms, parasites} --- ``Why do black people tend to not like any animals? Like it's just a bunny lol\ldots'' \\
\midrule
$-$ & 23 & \emph{Diasporic geolocation}: \textit{christchurch, saskatoon, chandigarh, queenstown, rotorua} --- ``How come? I just recently moved to Klang Valley from Penang\ldots'' \\
$-$ & 56 & \emph{Critical race reflexivity}: \textit{nuances, reflecting, subtleties, appreciating, dynamism} --- ``Being aware of histories of colonialism, racism, white supremacist attitudes is so important\ldots'' \\
$-$ & 21 & \emph{POC cultural achievement}: \textit{songwriter, acclaimed, showcased, singer-songwriter, performer} --- ``look at these beautiful talented black boys'' \\
\bottomrule
\end{tabular}
\caption{Cluster summary for the POC annotator conditional gradient ($\hat{g}_{\mathrm{low}}$). Positive pole: high hate-speech score; negative pole: counter-speech / low hate-speech score.}
\label{tab:glow_pos}
\end{table}

\subsection{White Annotators (\textit{g}\textsubscript{high})}

\subsubsection*{Gradient Summary}
The white annotator gradient produced a semantically differentiated positive
pole and a relatively compressed negative pole. The overall contrast ran
from lexically explicit, surface-marked hate speech---organized into four
distinct clusters at the high end---to a broadly affirmative but
undifferentiated counter-speech register at the low end. The gradient's
asymmetry between a finely parsed positive pole and a fused negative pole
is its most structurally distinctive feature relative to the general and
POC gradients.

\subsubsection*{Positive Pole}
The positive pole was parsed into four lexically distinct regions. One
cluster centered on eliminationist commands and explicit calls for violence
against ethnic and religious minorities. A second was organized around
racial and homophobic slurs combined with generalized contempt vocabulary.
A third captured the vermin and infestation register of dehumanizing
language. A fourth---absent from POC annotators' gradient as a standalone
region---reflected direct animal-species comparison applied to racial
groups. Taken together, these clusters suggest that white annotators apply
a more lexically granular schema to the \textit{dehumanize} sub-dimension
of \citet{kennedy2020constructing}, parsing zoological dehumanization
into the infestation metaphor and the species-comparison metaphor as
independent predictive signals rather than treating them as a unified
rhetorical mechanism.

\subsubsection*{Negative Pole}
The negative pole was organized around two clusters that merged semantic
material distinguished elsewhere. One large cluster fused celebratory
praise of POC achievement, reflective anti-racist commentary, and the
language of cultural dynamism into a single region---conflating semantic
sub-types that appear as separate clusters in POC annotators' gradient.
The second cluster was anchored by Anglo-North American civic geography,
with a lexicon of mid-size Canadian and American cities indexing ordinary
community and local identity discourse. Taken together, these clusters
suggest that lower hate speech scores for white annotators were associated
with broadly affirmative and non-threatening language, without the
structural differentiation between cultural celebration and deliberate
anti-racist reflexivity that characterizes the POC counter-speech pole.

\subsubsection*{Interpretation Note}
This corpus illustrates a pattern consistent with a threshold-based
detection schema: white annotators are sensitive to lexically explicit and
surface-marked hate speech signals---slurs, eliminationist vocabulary,
dehumanizing comparisons---but organize counter-speech as a residual
affirmative register rather than a structurally elaborated semantic domain.
The representative excerpts at the positive pole showed high cosine
alignment with unambiguous primary hate speech, and nearest-neighbor
structure and excerpt evidence were mutually reinforcing there. At the
negative pole, the fusion of distinct counter-speech registers into one
cluster, and the narrower geographic imagination it implies, is the primary
site of interpretive divergence from the general and POC gradients.

\begin{table}[htbp]
\centering
\small
\begin{tabular}{cc>{\raggedright\arraybackslash}p{0.68\linewidth}}
\toprule
Pole & Size & Summary (Top Words / Excerpt) \\
\midrule
$+$ & 30 & \emph{Eliminationist violence}: \textit{kill, attack, killing, destroy, enemies} --- ``@thegreatkhalid Muslims want to rape and destroy America. But okay.'' \\
$+$ & 22 & \emph{Slur contempt}: \textit{bastards, scum, n****r, f****t, criminals} --- ``Fuck: K***s S****s N*****s F*****s T*****s Commies\ldots'' \\
$+$ & 30 & \emph{Vermin infestation}: \textit{parasites, fleas, pests, cockroaches, infestation} --- ``This Islamic plague has to be eradicated!'' \\
$+$ & 18 & \emph{Species comparison}: \textit{pig, monkey, rabbit, squirrel, rat} --- ``Why do black people tend to not like any animals? Like it's just a bunny lol\ldots'' \\
\midrule
$-$ & 74 & \emph{Affirmative/celebratory}: \textit{passionate, phenomenal, showcased, nuances, dynamism} --- ``This is incredible, not only as an honor for them, but also as a sign of improving diversity\ldots'' \\
$-$ & 26 & \emph{Civic geography}: \textit{guelph, oakville, saskatoon, brampton, kitchener} --- ``How come? I just recently moved to Klang Valley from Penang\ldots'' \\
\bottomrule
\end{tabular}
\caption{Cluster summary for the white annotator conditional gradient ($\hat{g}_{\mathrm{high}}$). Positive pole: high hate-speech score; negative pole: counter-speech / low hate-speech score.}
\label{tab:ghigh_pos}
\end{table}

\subsection{Interaction Gradient (\textit{g}\textsubscript{int})}

\paragraph{Gradient summary.}
Unlike the main and conditional gradients, $\hat{g}_{\mathrm{int}}$ does
not describe what either annotator group finds hateful in absolute terms.
It describes where the meaning--rating association differs between
groups. Content that both groups rate similarly, whether highly hateful
or non-hateful, contributes little to this direction. The positive pole
therefore marks semantic neighborhoods where white annotators show
relatively stronger hate-speech sensitivity, whereas the negative pole
marks neighborhoods where POC annotators show relatively stronger
sensitivity. The small interaction-gradient norm
($\|\hat{g}_{\mathrm{int}}\| = 1.733$) confirms that this divergence is a
modest second-order signal over a largely shared response structure.

\paragraph{Positive pole.}
Three clusters loaded on the positive pole. The first and largest centered
on physical attack and violence vocabulary, including weapon references,
victimhood framing, and calls for lethal action. The second was a compact
cluster of biomedical and contagion vocabulary whose representative
excerpts indicate disease-based dehumanization directed at ethnic and
religious groups. The third consisted of meta-discursive racism
vocabulary: the language of accusation, bigotry, and ideological argument
about race. In the fitted interaction model, these neighborhoods mark
semantic regions where the association with higher hate-speech ratings
was relatively stronger for white annotators than for POC annotators.

\paragraph{Negative pole.}
Two clusters loaded on the negative pole. The first gathered vocabulary of
energy, vibrancy, and affect alongside content foregrounding racial
marginalisation and accountability. The second was anchored by
Spanish-language and Latinx community vocabulary; its representative
excerpts include intra-community conflict charged with colonial identity
politics and Spanish-language insults. In the fitted interaction model,
these neighborhoods mark semantic regions where the association with
higher hate-speech ratings was relatively stronger for POC annotators
than for white annotators.

\paragraph{Interpretation note.}
The interaction gradient is consistent with a modest asymmetry in the
semantic cues that marginally shift hate-speech ratings across annotator
groups. The positive pole is organized around relatively explicit or
surface-marked hostility, including attack vocabulary, biological
dehumanization, and overt racism discourse. The negative pole is organized
around more context-dependent material, including racialized affective
commentary and Spanish-language intra-community conflict. This pattern
suggests that some group-linked divergence may reflect differential access
to cultural or linguistic context, but the effect should not be
overstated: both groups share the dominant hate-speech gradient, and
$\hat{g}_{\mathrm{int}}$ captures only a smaller moderation component.

\begin{table}[htbp]
\centering
\small
\begin{tabular}{cc>{\raggedright\arraybackslash}p{0.68\linewidth}}
\toprule
Pole & Size & Summary (Top Words / Excerpt) \\
\midrule
$+$ & 46 & \emph{Attack and violence}: \textit{attack, killing, kill, victim, sniper} --- ``If these pathetic mayo's are only able to attack unarmed minorities we should give every minority their own assault rifle and vest.'' \\
$+$ & 11 & \emph{Contagion framing}: \textit{swine, meningitis, leprosy, allergy, bse} --- ``Whites tried to enslave Natives. Infectious diseases were the reason it didn't last long\ldots'' \\
$+$ & 43 & \emph{Meta-discursive racism}: \textit{racism, bigotry, racist, accusation, argument} --- ``If the liberals did not have white people to blame for racism, they would not have an excuse for being racists.'' \\
\midrule
$-$ & 64 & \emph{Affective racial commentary}: \textit{electrifying, gloriously, vibrancy, energizing, waning} --- ``America should be ashamed of the way it treats and marginalizes it's people of colour\ldots'' \\
$-$ & 36 & \emph{Spanish-language intra-community}: \textit{nuevo, nacional, m\'{e}xico, m\'{u}sica, barrio} --- ``For all you so called Puerto Ricans\ldots slap yourself with your grandmother's CHANCLA!!'' \\
\bottomrule
\end{tabular}
\caption{Cluster summary for the interaction gradient ($\hat{g}_{\mathrm{int}}$). Positive pole: themes where white annotators show amplified hate-speech sensitivity relative to POC annotators; negative pole: themes where POC annotators show amplified sensitivity.}
\label{tab:gint_pos}
\end{table}

\section{Formal Description of the Interaction SSD Model}
\label{appendix:formal}

Let $y_i$ denote the standardized outcome for observation $i = 1,\ldots,n$,
let $\mathbf{z}_i \in \mathbb{R}^K$ denote the PCA-reduced document vector
obtained from the SSD pipeline, and let $m_i$ denote the standardized
moderator. Binary moderators are first coded as $0/1$ from the relevant group
labels and then standardized.

The interaction SSD model is
\begin{equation*}
\begin{split}
  y_i &=
  \alpha
  + \mathbf{z}_i^\top \boldsymbol{\beta}
  + \gamma m_i \\
  &\quad + (\mathbf{z}_i m_i)^\top \boldsymbol{\delta}
  + \varepsilon_i ,
\end{split}
\end{equation*}
where $\boldsymbol{\beta} \in \mathbb{R}^K$ is the baseline semantic block,
$\gamma$ is the scalar moderator main effect, and
$\boldsymbol{\delta} \in \mathbb{R}^K$ is the semantic interaction block.
The effective semantic gradient in PCA space at moderator value $m^\ast$ is
therefore
\begin{equation*}
  \boldsymbol{\beta}(m^\ast)
  =
  \boldsymbol{\beta} + m^\ast \boldsymbol{\delta}.
\end{equation*}

Coefficients are stacked as
\[
\hat{\mathbf{h}}
=
[\alpha,\; \boldsymbol{\beta}^\top,\; \gamma,\;
\boldsymbol{\delta}^\top]^\top
\]
and estimated by OLS on the design matrix
\begin{equation*}
  \mathbf{X}
  =
  \bigl[
  \mathbf{1}_n
  \;\big|\;
  \mathbf{Z}
  \;\big|\;
  \mathbf{m}
  \;\big|\;
  \mathbf{Z} \odot \mathbf{m}
  \bigr],
\end{equation*}
where $\mathbf{Z}$ is the $n \times K$ matrix of PCA scores and
$\mathbf{Z} \odot \mathbf{m}$ denotes row-wise multiplication by the
moderator. The estimator and covariance matrix are
\begin{align*}
  \hat{\mathbf{h}}
  &=
  (\mathbf{X}^\top \mathbf{X})^{-1}
  \mathbf{X}^\top \mathbf{y}, \\
  \mathrm{Var}(\hat{\mathbf{h}})
  &=
  \hat{\sigma}^2
  (\mathbf{X}^\top \mathbf{X})^{-1}, \\
  \hat{\sigma}^2
  &=
  \frac{
  \|\mathbf{y} - \mathbf{X}\hat{\mathbf{h}}\|^2
  }{n - 2K - 2}.
\end{align*}

Because both $\boldsymbol{\beta}$ and $\boldsymbol{\delta}$ are
$K$-dimensional coefficient blocks, their significance is assessed with Wald
$F$ tests. For a block $\mathbf{c}$ of length $q$ with covariance submatrix
$\mathbf{V}_{cc}$,
\begin{equation*}
\begin{split}
  F_{\mathrm{Wald}}
  &=
  \frac{\mathbf{c}^\top \mathbf{V}_{cc}^{-1} \mathbf{c}}{q} \\
  &\sim
  F(q,\; n - 2K - 2)
  \quad \text{under } H_0: \mathbf{c}=\mathbf{0}.
\end{split}
\end{equation*}
The test on $\boldsymbol{\delta}$ is equivalent to an incremental test
comparing the full model against a reduced model that retains the semantic
block and moderator main effect but omits the interaction block. The scalar
moderator effect $\gamma$ is reported with its standard error, $t$ statistic,
and two-tailed $p$ value.

To recover embedding-space gradients, the PCA-space coefficients are
back-projected through the PCA loading matrix and rescaled by the pre-PCA
feature standard deviations. Let
$\mathbf{P} \in \mathbb{R}^{D \times K}$ denote the PCA loading matrix and
$\mathbf{s} \in \mathbb{R}^D$ the vector of feature standard deviations from
the pre-PCA standardization step. Then
\begin{align*}
  \hat{g}_{\mathrm{main}}
  &=
  (\mathbf{P}\boldsymbol{\beta}) \oslash \mathbf{s}, \\
  \hat{g}_{\mathrm{int}}
  &=
  (\mathbf{P}\boldsymbol{\delta}) \oslash \mathbf{s}, \\
  \hat{g}(m^\ast)
  &=
  \hat{g}_{\mathrm{main}}
  +
  m^\ast \hat{g}_{\mathrm{int}} .
\end{align*}
For binary moderators, $m^\ast$ corresponds to the standardized values of the
two coded groups; for continuous moderators, it can be set to theoretically
meaningful values such as one standard deviation below and above the mean.

Each recovered gradient can be interpreted with the standard SSD procedure:
nearest neighbors are retrieved around its positive and negative poles,
clustered separately, and linked back to representative corpus snippets.
The $\boldsymbol{\delta}$-block Wald test, rather than the magnitude
$\|\hat{g}_{\mathrm{int}}\|$, should be treated as the primary test of
moderation. If this test is non-significant, the interaction gradient and
conditional-gradient differences should be interpreted descriptively rather
than confirmatorily.

\section{Reanalysis with the Common Crawl 840B GloVe Model}
\label{app:cc840b}

To probe robustness to embedding choice, we re-ran the full interaction-SSD pipeline on
the same corpus using the 300-dimensional Common Crawl 840B GloVe model
\citep{pennington2014glove}, with all preprocessing and sweep settings held identical.
The sweep selected $K = 52$ components.

Table~\ref{tab:dolma-regression} summarises the regression results. All three blocks
replicated: the semantic block ($F(52, 44708) = 1308.64$, $p < 10^{-16}$), the race
main effect ($\hat{\gamma} = .024$, $t(44708) = 7.28$, $p = 3.36 \times 10^{-13}$),
and the interaction block ($F(52, 44708) = 15.18$, $p < 10^{-16}$). The interaction
effect again remained small in variance-explained terms, with partial $R^2 = .017$,
closely matching the 42B result ($.019$). Thus, the moderation signal was robust to
embedding choice while retaining the same substantive interpretation: annotator race
sensitivity appeared as a reliable but modest second-order deviation from the dominant
shared semantic rating structure.

\begin{table}[htbp]
\centering
\small
\renewcommand{\arraystretch}{1.15}
\setlength{\tabcolsep}{3.5pt}
\resizebox{\linewidth}{!}{%
\begin{tabular}{lcccc}
\toprule
 & \textbf{Overall} & \textbf{Semantic} & \textbf{Race} & \textbf{Interaction} \\
 & \textbf{model} & \textbf{block} & \textbf{main} & \textbf{block} \\
\midrule
Test          & $F$      & $F$       & $t$      & $F$ \\
Statistic     & $736.38$ & $1308.64$ & $7.28$   & $15.18$ \\
df            & $105,44708$ & $52,44708$ & $44708$ & $52,44708$ \\
$p$           & $<10^{-16}$ & $<10^{-16}$ & $3.36{\times}10^{-13}$ & $<10^{-16}$ \\
\midrule
$R^2$         & $.633$   & ---       & ---      & --- \\
Estimate      & ---      & ---       & $.024$   & --- \\
Partial $R^2$ & ---      & ---       & ---      & $.017$ \\
\bottomrule
\end{tabular}
}
\caption{Interaction-SSD regression results with the Common Crawl 840B GloVe
model. $n = 44{,}814$; outcome and IV were $z$-scored; $K = 52$. Partial $R^2$
is reported for the interaction block and represents the proportion of residual
variance explained by the interaction after accounting for the shared semantic
rating structure and the race main effect.}
\label{tab:dolma-regression}
\end{table}

To interpret $\hat{g}_{\mathrm{main}}$, we clustered nearest neighbors on both poles.
Table~\ref{tab:dolma-gmain} summarises the resulting themes. The positive pole reproduced
the three substantive clusters from the main analysis (slur contempt, violent/eliminationist
vocabulary, vermin/infestation framing), with a fourth cluster of bibliographic tokens
(\textit{j., sci., appl, lett}) appearing as a noise artefact with markedly lower coherence.
The negative pole was structurally richer, matching the main analysis on diasporic geolocation
and reflexive cultural discourse while adding a POC professional-and-advocacy cluster
(interpreters, narrators, liaison vocabulary) and a performing-arts cluster (soloists,
choral, orchestral).

\begin{table}[!t]
\centering
\small
\begin{tabular}{cc>{\raggedright\arraybackslash}p{0.68\linewidth}}
\toprule
Pole & Size & Summary (Top Words / Excerpt) \\
\midrule
$+$ & 23 & \emph{Slur contempt}: \textit{fuckers, bastards, scum, murderous, scumbag} --- ``fuck israel. fucking scumbags.'' \\
$+$ & 29 & \emph{Violent/eliminationist}: \textit{kill, trap, rip, killing, destroy} --- ``Fuck Serbians! Fucking imbecile cunts, never forget the genocides and murders they did.'' \\
$+$ & 33 & \emph{Vermin}: \textit{parasites, rodents, pests, ants, insect} --- ``The usa does not need any more rats and cockroaches, get them out!'' \\
$+$ & 15 & \emph{[Noise]}: \textit{j., j, sci., appl, lett} --- diverse low-coherence excerpts unrelated to hate speech \\
\midrule
$-$ & 25 & \emph{POC professional/advocacy}: \textit{interpreters, narrators, liaising, midwives, attuned} --- ``Essential resources for women and PoC in the academy!'' \\
$-$ & 21 & \emph{Diasporic geolocation}: \textit{midlands, kelowna, kamloops, nanaimo, cumbria} --- ``How come? I just recently moved to Klang Valley from Penang\ldots'' \\
$-$ & 40 & \emph{Reflexive cultural discourse}: \textit{dynamism, immediacy, authenticity, artistry, emphasising} --- ``This is incredible, not only as an honor for them, but also as a sign of improving diversity\ldots'' \\
$-$ & 14 & \emph{Performing arts}: \textit{soloists, soloist, choral, operatic, orchestral} --- ``Not to take away from her as she did amazing. But there has been quite a few black opera singers brfore.'' \\
\bottomrule
\end{tabular}
\caption{Cluster summary for $\hat{g}_{\mathrm{main}}$ under the CC840B model.
Positive pole: high hate-speech score; negative pole: counter-speech / low hate-speech score.}
\label{tab:dolma-gmain}
\end{table}

The interaction gradient $\hat{g}_{\mathrm{int}}$ showed a different semantic organisation
(Table~\ref{tab:dolma-gint}): its positive pole organised around epistemic and argumentative
discourse (reasoning vocabulary, academic credentials, clinical framing) and its negative
pole around consumer and entertainment contexts (automobile showrooms, film premieres,
consumer technology) alongside Spanish-language intra-community content---in contrast to
the attack vocabulary, contagion framing, and meta-discursive racism that anchored the
42B $\hat{g}_{\mathrm{int}}$.

Because the $F$-test on the interaction block is estimated in PCA space without reference
to neighbourhood structure, it is robust to this variation; the semantic interpretation of
gradient poles, by contrast, depends on the local geometry of the embedding model.
Both models agree that annotator racial identity moderates the hate-speech semantics, but
disagree on which surface cues carry that moderation signal in their respective embedding
spaces.

\clearpage
\begin{table}[!ht]
\centering
\small
\begin{tabular}{cc>{\raggedright\arraybackslash}p{0.68\linewidth}}
\toprule
Pole & Size & Summary (Top Words / Excerpt) \\
\midrule
$+$ & 25 & \emph{Epistemic/argumentative}: \textit{arguments, reasoning, evidence, conclusions, rationale} --- ``Ben Shapiro Debunks Black Lives Matter with FACTS and LOGIC.'' \\
$+$ & 37 & \emph{Academic credentials}: \textit{professor, phd, researcher, dr., scientist} --- ``34 Black women are graduating from West Point in 2019, making it the most diverse class in the academy's history'' \\
$+$ & 38 & \emph{Clinical/medical framing}: \textit{clinical, disease, study, diseases, findings} --- ``Asian-white couples face distinct risks during pregnancy, new study reveals'' \\
\midrule
$-$ & 29 & \emph{Consumer/automobile}: \textit{showroom, showrooms, regal, sedans, cherokee} --- ``Trump will have a gaggle of Black men carry him up the stairs of the Lincoln memorial in a sedan chair.'' \\
$-$ & 38 & \emph{Entertainment representation}: \textit{debuts, debuted, premiere, premieres, gala} --- ``@crissles and when I saw the Broadway touring version Ariel was Asian.'' \\
$-$ & 12 & \emph{Consumer technology}: \textit{chromecast, roku, pre-orders, handsets, preorders} --- ``What is happening to btsworld i cant open the app'' \\
$-$ & 21 & \emph{Spanish-language intra-community}: \textit{nuevo, ciudad, desde, tierra, norte} --- ``For all you so called Puerto Ricans\ldots slap yourself with your grandmother's CHANCLA!!'' \\
\bottomrule
\end{tabular}
\caption{Cluster summary for $\hat{g}_{\mathrm{int}}$ under the CC840B model.
Positive pole: themes where white annotators show amplified hate-speech sensitivity;
negative pole: themes where POC annotators show amplified sensitivity.}
\label{tab:dolma-gint}
\end{table}

\end{document}